\documentclass[review]{elsarticle}

\usepackage{lineno,hyperref}
\modulolinenumbers[5]
\usepackage{subfig}
\usepackage{array}
\usepackage{multirow}
\usepackage{bbm}
\usepackage{amssymb}

\newtheorem{thm}{Theorem}
\usepackage{rotating}
\usepackage{adjustbox}

\journal{Journal of Information Fusion}

\begin{document}

\begin{frontmatter}
\title{A novel multimodal fusion network based on a joint coding model for lane line segmentation}

%
\author[address1,address2]{Zhenhong Zou}
\ead{zhenhongzzh@gmail.com}
\author[address1,address2]{Xinyu Zhang\corref{mycorrespondingauthor}}
\cortext[mycorrespondingauthor]{Corresponding author}
\ead{xyzhang@tsinghua.edu.cn}
\author[address3]{Huaping Liu}
\ead{hpliu@tsinghua.edu.cn}
\author[address1,address2]{Zhiwei Li}
\ead{lizhiwei713818@163.com}
\author[address4]{Amir Hussain}
\ead{hussain.doctor@gmail.com}
\author[address1,address2]{Jun Li}
\ead{lijun1958@tsinghua.edu.cn}

\address[address1]{State Key Laboratory of Automotive Safety and Energy, Tsinghua University, Beijing, China}
\address[address2]{School of Vehicle and Mobility, Tsinghua University, Beijing, China}
\address[address3]{Department of Computer Science and Technology, Tsinghua University, Beijing, China}
\address[address4]{Edinburgh Napier University, U.K.}

\begin{abstract}

There has recently been growing interest in utilizing multimodal sensors to achieve robust lane line segmentation. In this paper, we introduce a novel multimodal fusion architecture from an information theory perspective, and demonstrate its practical utility using Light Detection and Ranging (LiDAR) camera fusion networks. In particular, we develop, for the first time, a multimodal fusion network as a joint coding model, where each single node, layer, and pipeline is represented as a channel. The forward propagation is thus equal to the information transmission in the channels. Then, we can qualitatively and quantitatively analyze the effect of different fusion approaches. We argue the optimal fusion architecture is related to the essential capacity and its allocation based on the source and channel. To test this multimodal fusion hypothesis, we progressively determine a series of multimodal models based on the proposed fusion methods and evaluate them on the KITTI and the A2D2 datasets. Our optimal fusion network achieves 85\%+ lane line accuracy and 98.7\%+ overall. The performance gap among the models will inform continuing future research into development of optimal fusion algorithms for the deep multimodal learning community.

\end{abstract}

\begin{keyword}
Multimodal fusion\sep Information theory\sep Lane line segmentation\sep Semantic segmentation\sep Neural Network

\end{keyword}

\end{frontmatter}


\section{Introduction}

In autonomous driving, lane line detection can indicate the driving area and direction by identifying the marks on the ground. In this paper, we focus on its first step, lane line segmentation, which is the key for the subsequent curve fitting.

\par Current methods can be divided into three types: camera-based, LiDAR-based and fusion methods. Camera images contain abundant information and have been proved to be effective in segmentation tasks. But they would not work well under weak or changing illumination due to the limited photosensitivity of cameras. Previous research has explored possible solutions by utilizing specific features, like the slender shape of lane lines\cite{pan2018spatial,he2016accurate} and relationships among the continuous images\cite{zou2019robust}. Besides, LiDAR is a practical alternative for it provides 3D point clouds around the vehicle with information reflecting the materials and shapes of the obstacles. Different from cameras, LiDARs can avoid the influence of lights, while the sparsity of point clouds limits the resolution in perception.
\par As a complement, fusion methods can leverage advantages from multimodal data\cite{caltagirone2019lidar,bai2018deep}. They can fuse the source data, feature maps or model output at different stages of the models to obtain comprehensive patterns of the target. However, though the latest research has presented leading results, few of them provide reliable illustration on the fusion mechanism, or common rules in fusion model design\cite{feng2019deep}.
\par To build an interpretable fusion model, we proposed to review multimodal fusion with information theory. As most of the deep networks can be regarded as cascaded feature extraction layers, we formulate each single node, layer and pipeline, even the whole network as a communication channel\cite{mackay2003information2,Tishby2015}. Therefore, based on the inference of the Shannon's Theorems, the channel structure and information source determine the upper and lower bounds (channel capacity and rate-distortion bound) of the network learning ability. Then we suggest the balance about capacity and its distribution accordingly.

\begin{figure}[t]
\centering
  \begin{tabular}{@{}c@{ }c@{ }c@{ }c@{ }c@{}}
    \subfloat[Ground Truth]{%
  \begin{tabular}{@{}c@{}}
  \includegraphics[width=0.19\textwidth]{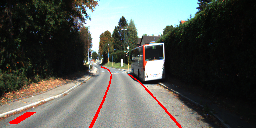}\\
    \includegraphics[width=0.19\textwidth]{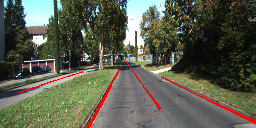}\\
   \includegraphics[width=0.19\textwidth]{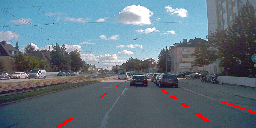}\\
    \includegraphics[width=0.19\textwidth]{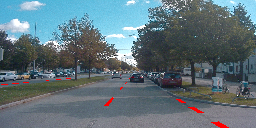}
  \end{tabular}}
    &
      \subfloat[LaneNet]{%
    \begin{tabular}{@{}c@{}}
	\includegraphics[width=0.19\textwidth]{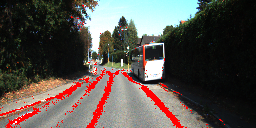}\\
	\includegraphics[width=0.19\textwidth]{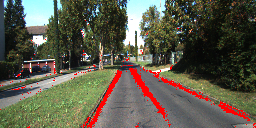}\\
	\includegraphics[width=0.19\textwidth]{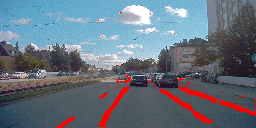}\\
	\includegraphics[width=0.19\textwidth]{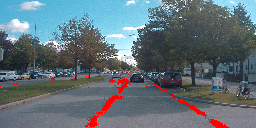}
    \end{tabular}}
 &
   \subfloat[SCNN]{%
 \begin{tabular}{@{}c@{}}
 \includegraphics[width=0.19\textwidth]{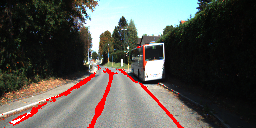}\\
 \includegraphics[width=0.19\textwidth]{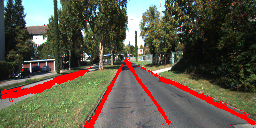}\\
 \includegraphics[width=0.19\textwidth]{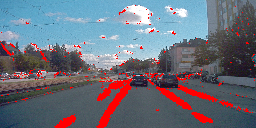}\\
 \includegraphics[width=0.19\textwidth]{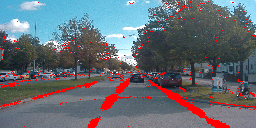}
 \end{tabular}}
	 &
	   \subfloat[Baseline]{%
	 \begin{tabular}{@{}c@{}}
	 \includegraphics[width=0.19\textwidth]{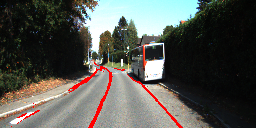}\\
	 \includegraphics[width=0.19\textwidth]{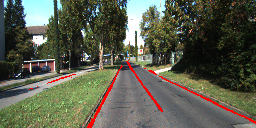}\\
	 \includegraphics[width=0.19\textwidth]{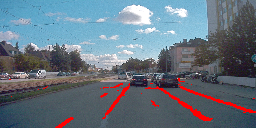}\\
	 \includegraphics[width=0.19\textwidth]{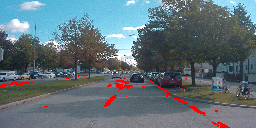}
	 \end{tabular}}
	 &
	   \subfloat[Final Model]{%
	 \begin{tabular}{@{}c@{}}
	 \includegraphics[width=0.19\textwidth]{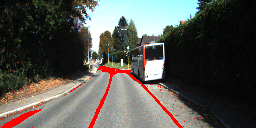}\\
	 \includegraphics[width=0.19\textwidth]{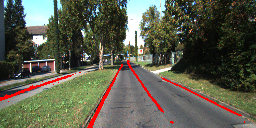} \\
	 \includegraphics[width=0.19\textwidth]{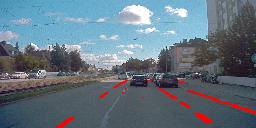} \\
	 \includegraphics[width=0.19\textwidth]{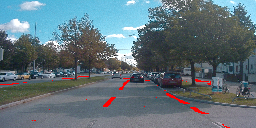}
	 \end{tabular}}
	
  \end{tabular}
  \caption{Results of Two State-of-the-art Models(SCNN\cite{pan2018spatial}, LaneNet\cite{neven2018towards}) And Ours: the first two rows are tested on the KITTI dataset and the rest are tested on the A2D2 dataset.}
  \label{CompareResult}
\end{figure}

\par In addition, we also conduct comparative experiments to verify our inference, and introduce a novel fusion network derived from them. On one hand, we explore the respective effect of different factors in fusion, including the fusion stage and network structure. On the other hand, we test the influence from the information source via modality lost experiments. According to the results on the KITTI\cite{Geiger2012CVPR} and A2D2\cite{geyer2020a2d2} benchmark, our optimal model gain 8.51\% lane line accuracy or 7.6 F2-score at most compared with the baseline. It also performs more robustly with a single modality lost. As shown in the Fig.~\ref{CompareResult}, the result demonstrates the benefits of our fusion strategy. Additionally, our lightweight networks are built end-to-end that can achieve real-time segmentation at over 64.9 FPS with 85\%+ lane line accuracy and 98.7\%+ overall on both datasets.
\par In conclusion, our main contribution are:
\begin{itemize}
\item We propose to review the multimodal network from the perspective of channel model and provide reasonable illustration for fusion.
\item We provide qualitative and quantitative analysis on different factors that can influence the fusion networks based on information theory.
\item We put forward a novel multimodal fusion network, which can approach the cutting-edge performance on lane line segmentation tasks.
\end{itemize}
\par In the following sections, we first review the recent progress in lane line segmentation, multimodal fusion and information theory. Then, we illustrate how to model and analyze the network as in the framework of channel model, and discuss the fusion strategy referring to Shannon's theory. Moreover, we will conduct quantitative analysis on the channel and source in the experiments and ablation study.

\section{Related Work}
\subsection{Image-based Lane Line Segmentation}
In 2015, Brody et al. applied convolutional neural networks (CNN) to lane line detection (segmentation) tasks\cite{huval2015empirical} and led the trend of CNN\cite{he2016accurate}. At the same time, end-to-end training became common. For example, the LaneNet proposed by Davy et al.\cite{neven2018towards} and the SCNN by X. Pan et al.\cite{pan2018spatial} were built end-to-end. These models utilized the geometric information of the lane line strips, but some researchers also considered the generative model, like the EL-GAN\cite{ghafoorian2018gan}. Two novel methods were proposed in 2019: the ConvLSTM\cite{zou2019robust} can leverage the association information of continuous frame images by combining the LSTM and U-Net; the ENet-SAD\cite{hou2019learning} improves the effect by using the self-attention model. In conclusion, most approaches can be summarized in four types: geometric models, recurrent networks, generative models and attention models.
\par However, some problems remain unsolved, such as the changing illumination for camera-base methods. In 2017, S. Lee et al. proposed the VPGNet to study detection under a variety of weak lighting conditions, such as rain and fog weather, and shadowy ground. The ConvLSTM\cite{zou2019robust} attempts to predict lane lines using contextual information. Although they can improve the performance in some cases, they cannot handle the problem completely.

\subsection{Multimodal Fusion}
Multimodal learning was put forward to combine the advantages and deal with the shortcomings in different data in various fields\cite{meng2020survey,khaleghi2013multisensor}, from affective computing\cite{poria2017review} to autonomous driving\cite{feng2019deep}. The research\cite{poria2017review,gogate2017deep} shows that textual, vocal and visual data can provide cues from multiple aspects to better identify psychological patterns, which also leads to the high-dimension feature assumption and the motivation of fusion in perception. Data fusion potentially achieve better performance, among the area of cross-modality detection\cite{zhang2019cross,schlosser2016fusing} and tracking\cite{cho2014multi}, 3D object detection\cite{ku2018joint,chen2017multi} and deception detection\cite{gogate2017deep}. Zhang et al.\cite{zhang2019cross} utilize cross-modality interactive attention among multispectral input like RGB and thermal images and further inform the importance of adapative fusion strategy. Instead, Joel's method\cite{schlosser2016fusing} detects pedestrian with LiDAR and camera. Hyunggi\cite{cho2014multi} proposed a method to combine Radar, LiDAR and camera for vehicle detection and tracking. The AVOD\cite{ku2018joint} and MV3D\cite{chen2017multi} were proposed to utilize spatial information in 3D detection. Current studies in deep learning mainly can be divided into  three types of fusion: early fusion, middle fusion and late fusion. Early fusion combines multimodal data in preprocess, while middle fusion is for the feature processing stage and late fusion focuses on the output from several
pipelines to generate the final result.

\par Though various fusion methods have been proposed, few of them are targeted on lane line segmentation. To achieve robust model, the point clouds reflectance map and images can be fused to combine different patterns. M. Bai et al.\cite{bai2018deep} propose a deep multi-sensor detection network that applies CNN to generate a dense pseudo image from point clouds, which is projected to bird-eye-view plane with an RGB image. L. Caltagirone et al.\cite{caltagirone2019lidar} fuse the feature maps of point clouds and RGB images in 20 convolution layers for adaptive fusion. Though previous research shows that feature fusion performs better than others, the optimal fusion stage and the fusion mechanism are still unclear\cite{feng2019deep,schlosser2016fusing,liang2019multi,guan2019fusion}.
For further information, we refer readers to the survey\cite{feng2019deep,meng2020survey} that reviews the fusion methods in machine learning and autonomous driving proposed in recent years.

\subsection{Information Theory}
Information theory, proposed by C. Shannon\cite{shannon1962the}, studies the information process based on the quantification of information with the concept of entropy. By quantifying the amount of uncertainty in signals, the information entropy reveals the limits on signal processing and communication operations. Other important measures in information theory include mutual information, channel capacity, and error exponents, etc.\cite{information2000} However, they seldom appear in fusion network research. MacKay et al. have discussed modeling a neuron or a network as a channel\cite{mackay2003information2}. N. Tishby et al. proposed the information bottleneck theory (IB) with a variational principle to reflect on signal processing problems\cite{tishby2001the}. In 2015, Tishby and Zaslavsky further revealed the mechanism deep learning model using the IB\cite{Tishby2015}. The experiments on a multi-layer perceptron shows that, the network tends to capture related information first and comprise them later. Though the result has not been generalized well to more complex CNN models, it inspires us to review the fusion model as a channel and the fusion process as information gain.

\section{Multimodal Fusion from The Perspective of Information Theory}
To reveal the relationship between the information theory and multimodal fusion, we first model the network as a joint coding channel, then it will be able to uncover the mechanism of fusion with Shannon's Theorems. Finally, we will introduce some practical fusion strategy derived from the analysis.

\subsection{Network as A Joint Coding Channel}
\begin{figure}[t]
\centering
  \begin{tabular}{c}
    \subfloat[The Communication System in Shannon's Paper\cite{shannon1962the}]{%
  \includegraphics[width=0.8\textwidth]{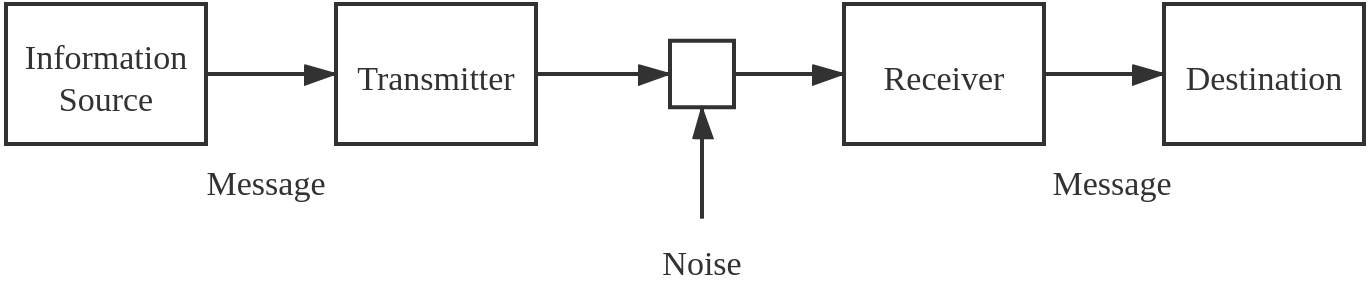}}\\
    \subfloat[Model The Fusion Network as A Joint Coding Channel]{%
  \includegraphics[width=1\textwidth]{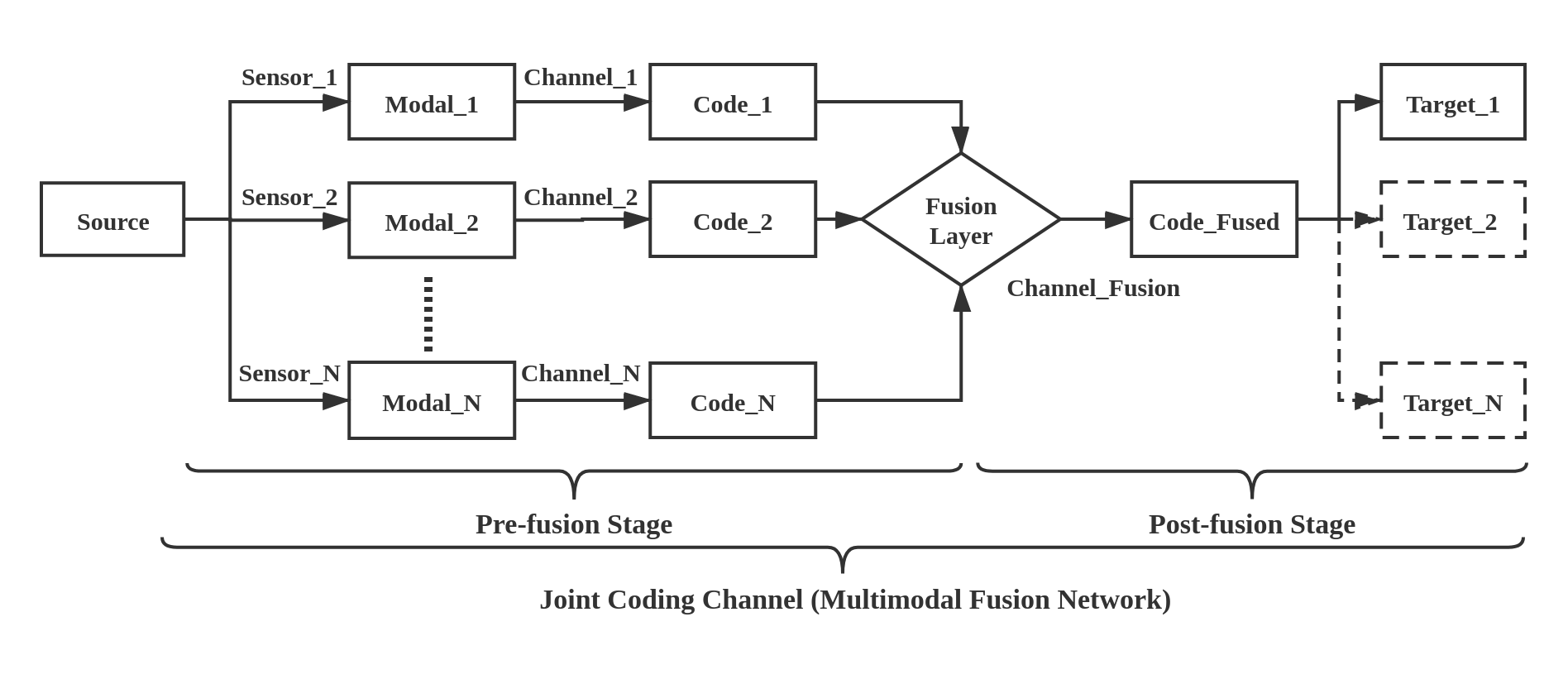}}
\end{tabular}
\caption{The Traditional Communication System And The Multimodal Network: for fusion model, different modal data are processed in individual channels, fused in the fusion channels and decoded to be the desired output, which is the projection of the target in one modality.}
\label{communicationmodel}
\end{figure}

As shown in the Fig.~\ref{communicationmodel}, a basic communication system includes five parts: information source, transmitter, channel, receiver and destination. To formulate the deep network as a channel, the five parts become: the source to measure, the encoder (including sensors and encoders), the hidden layers (channel), the decoder and the desired output. Generally, we can also view the learning as joint source-channel coding, in which the encoder, channel and decoder are combined. In this way, we can further look into every single layer for more precise analysis on the fusion. Specifically, each layer or pipeline equals to a channel, and the whole network becomes a cascaded channel.
\par To make it clear, we hereby define that: the output channel number in each layer equals to the code length $n$, the features equal to the codes, and the learning ability equals to the channel capacity. The code set $M$ therefore contains all the possible features, and the rate of code $R=\frac{log_2|M|}{n}$ represents the efficiency of the code, that is, the redundancy of network. Researchers have also discussed the estimation of the information entropy and mutual information\cite{tishby2001the,gutmann2010noise,belghazi2018mutual}, but they have not developed an optimal method for all networks. Thus, we assume that the networks we mentioned have a potential entropy, which is not used directly in the following. Though we ignore many details in the definition, we can conduct basic qualitative analysis and quantitative comparison for the fusion problems in multimodal lane line segmentation. 
\par In addition, in the case of multimodal fusion network, the fused features and information are generated in the joint coding in the fusion channels. As the multimodal data and features in the coding comes from the individual $sensor\_i$ and corresponding $channel\_i (i\in\{1,2,...,N\})$, multimodal learning is actually a kind of parallel joint coding.

\subsection{Shannon's Theorems in Multimodal Learning}
As the fundamentals of the information theory, Shannon's Theorems reveal the connection between coding and information transmission ability for certain source and channel models. They also imply that the optimal multimodal fusion architecture is driven by both the data quality and the network itself. First, we briefly recall Shannon's Theorems as below\cite{shannon1962the,information2000}:
\begin{thm}[Noisy-Channel Coding Theorem] Let R be the rate of code, and C be the channel capacity. For any discrete memoryless channel, if $R<C$, then R is achievable. Conversely, if $R>C$, it is not achievable.
\label{th2}
\end{thm}
\begin{thm}[Shannon-Hartley Theorem]
The channel capacity $C$ is:
\begin{eqnarray}\label{12}
C = B \times log(1+S/N)
\end{eqnarray}
where $B$ is the bandwidth of channels and $S/N$ is the signal-noise ratio(SNR).
\label{th3}
\end{thm}
\begin{thm}[Rate-Distortion Theorem]
The rate distortion function for an i.i.d. source X with distribution $p(x)$ and bounded distortion function $d(x,\hat{x})$ is equal to the associated information rate distortion function. Thus
\begin{equation}
R(D)=\min_{p(\hat{x}|x):\Sigma_{(x,\hat{x})p(x)p(\hat{x}|x)d(x,\hat{x})}\leq D}I(X;\hat{X})
\end{equation}
is the minimum achievable rate at distortion D.
\label{th4}
\end{thm}
\par In summary, these theorems focus on the source coding and the transmission in channels. The Theorem.~\ref{th2} and \ref{th3} indicate that the rate of code is bounded by the channel capacity, which is determined by channel bandwidth and SNR. The Theorem.~\ref{th4} provides a lower bound for the rate. Notice that the channel capacity only reflects the performance of channel, while rate-distortion targets the source. By combining them we found that for the given distortion $D$, a code is achievable if and only if the code, channel and source satisfy the inequality:
\begin{equation}
R(D)\leq R\leq C
\label{conflict}
\end{equation}

\paragraph{Single-modal Learning} These theorems can be directly transferred to deep learning, which present the limits of learning (channel capacity) and compressing (rate-distortion). To avoid the complicated calculation of network capacity like MacKey et al.\cite{mackay2003information2} and simplify the problem, we will perform qualitative analysis on the model instead. Specifically, for single-modal learning, the rate-distortion function has been determined based on the dataset, and the channel capacity can be adjusted automatically during training or manually. Moreover, we can rewrite the Eq.~\ref{conflict} informally as below:
\begin{equation}
D=D(R)\geq D(C)
\label{newconflict}
\end{equation}
\par The Eq.~\ref{newconflict} shows that by properly raising the capacity and rate of code, the network can achieve less distortion. But the improvement is finited for the limited source information.

\paragraph{Multimodal Learning} The case in multimodal learning will be slightly different from the single-modal, especially the channel capacity and source distortion. Assume that the object has a high-dimension feature space, and different sensors only observe its subspace in specific modalities. As shown in the Fig.~\ref{communicationmodel}, a basic fusion model can be separated into two stages: pre-fusion and post-fusion. In pre-fusion, it holds the inference for single-modal learning in each pipeline. But in post-fusion, the input is the combination of feature subspace, and the $channel\_fused$ is the extension of sub-channels in pre-fusion. That indicates the individual rate-distortion function and required capacity for each modality, which will further influence the optimal bandwidth (weight) allocation of the sub-channels in fusion, with the balance between pre-fusion and post-fusion. In addition, the capacity is determined by the bandwidth and SNR. The bandwidth measures the size of feature subspace, and SNR reflects the ratio of effective information transmitted through the network. The SNR of data means the SNR in sensor measurement, which is also a channel.
\par In conclusion, the channel model and Shannon's Theorems provide a novel perspective to review the deep multimodal learning. Furthermore, they also imply that the optimal fusion architecture depends on three balance in: the capacity allocation for sub-channels, the division of pre-fusion and post-fusion, the conflict in information distortion and channel redundancy. A fusion model is essentially approaching the balance by improving the channel capacity or using a better information source. Then it will be able to develop novel information-driven fusion models with our deduction.

\subsection{Information-driven Fusion Strategy} 
Towards the optimal LiDAR-camera fusion in lane line segmentation, we propose a novel information-driven multimodal fusion strategy with a two-stage analysis on the information source and network. 
\par First, we estimate the required channel capacity based on the information source, including the capacity in the overall model and its allocation for each modality. Generally, uncertain data (with large entropy) transmit more message in the channel and require higher rate of code, namely more capacity according to Theorem.~\ref{th2}. Supplementally, the divergence between the training set and test set also add to the uncertainty. It is the same in the case of capacity allocation for the pre-fusion pipelines. Those data with higher SNR, for example a well-captured image compared with the sparse point clouds, is supposed to occupy larger capacity in fusion. 
\par Besides, the optimal fusion stage also depends on the data SNR. Actually, as the bandwidth decreases during fusion process, post-fusion forces the model to 'forget' noise in joint optimization and enhance the generalization to other tasks. Conversely, using more pre-fusion would help fitting the training set for it adopts more information from the source. Therefore, when the uncertainty increases, like using small training set or hard test set, ealier fusion will perform better.
\par Then, we can adjust the channel architecture with these estimation, and all of them are essentially changing the capacity by bandwidth or SNR. For the capacity allocation, it is practical to apply adaptive bandwidth/weight in fusion, for instance the adaptive weight estimator, fully linear connection, depth-wise convolution or other individual process for different modal data. But for the overall capacity, we consider using cascaded channel structure, including tandem channels (like attention model and cascaded detectors) and parallel channels (like the Y-shaped fusion in Fig.~\ref{communicationmodel}). Both types aim at increasing the capacity with longer codes: the tandem focus on the depth of network, and the parallel target the width. As the optimal model can be the combination of them, we integrate the road segmentation with our model as multi-task learning, and combine the early and middle fusion to extend the capacity, which comprise both tandem and parallel structure. Details are referred to the model section.
\par Another practical theorem is about error correction\cite{information2000}:
\begin{thm}
A $q$-ray $(n,|M|,d)$ code $C$ can correct $t$ errors if and only if:
\begin{eqnarray}
d(C)\geq 2t+1
\end{eqnarray}
where $n$ is the code length and $d(C)$ is the minimal distance among codes.
\label{th5}
\end{thm}
\par The theorem shows that a model can obtain better representation of the source with longer code, but it will perform less robustly. It not only informs the conflict between essential capacity and redundancy, but also indicates a possible approach to compare the contribution of different modalities by computing their error correction ability.
\par Beyond these, although the qualitative analysis method has been proposed above, it is unable to direct the precise fusion model without quantification tools. Instead, it provides guidance for the architecture. Specific fusion models will be introduced in the following sections.

\section{Multimodal Models}
To better demonstrate our study, we conduct a series of experiments to explore the effect of different fusion stages and methods. We first introduce our baseline model, then turn to the multimodal fusion approaches.
\subsection{Single Modal Network}
\begin{figure*}[t]
\centering
\includegraphics[width=1\textwidth]{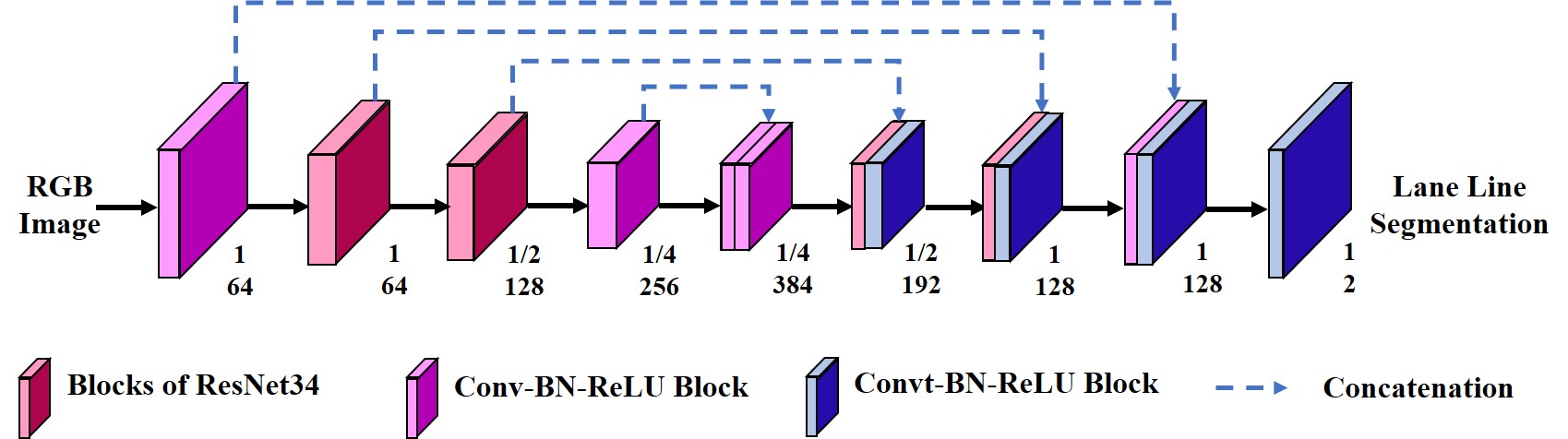}
\caption{Our Single Modal Baseline Model (V1): it takes an RGB image as input and outputs a 256x128 binary map.}
\label{baseline}
\end{figure*}

Considering lane line segmentation is a class-unbalanced task, we build up our baseline model based on the U-Net\cite{unet2015}. As shown in Fig.\ref{baseline}, it comprises 4 blocks in the encoder and 5 in the decoder, in which two are ResNet-34 blocks, the last four layers use transposed convolution and the rest are convolutional blocks. All convolutional blocks have a batch-normalization layer and a ReLU layer following the convolution layer, and all kernels size are 3x3. Each block in the encoder is linked to the corresponding blocks in the decoder with a dash line, that concatenate the output of them to correct the feature maps.

\begin{figure*}[!htp]
\centering
  \begin{tabular}{@{}c@{}}
  \subfloat[V2\&V3: Feature fusion at the early stage.]{%
    \includegraphics[width=1\textwidth]{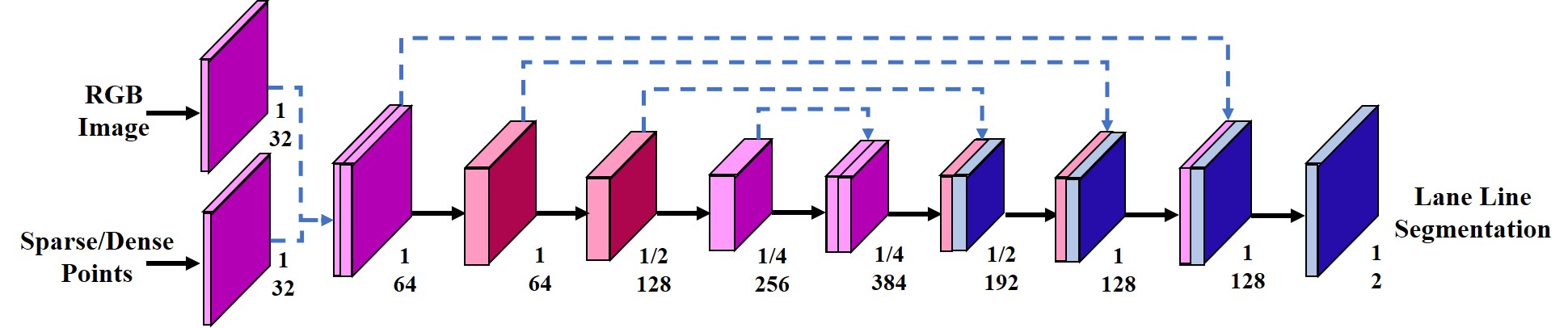}}\\
  \subfloat[V4: Feature fusion at the middle stage.]{%
    \includegraphics[width=1\textwidth]{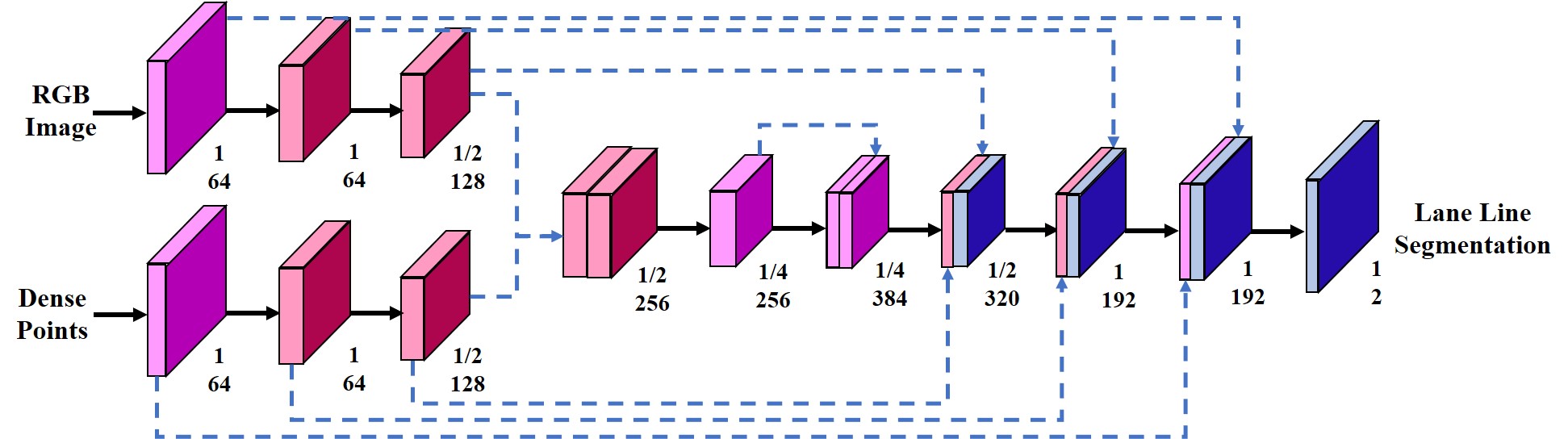}} \\
  \subfloat[V5: Feature fusion at the late stage.]{%
    \includegraphics[width=1\textwidth]{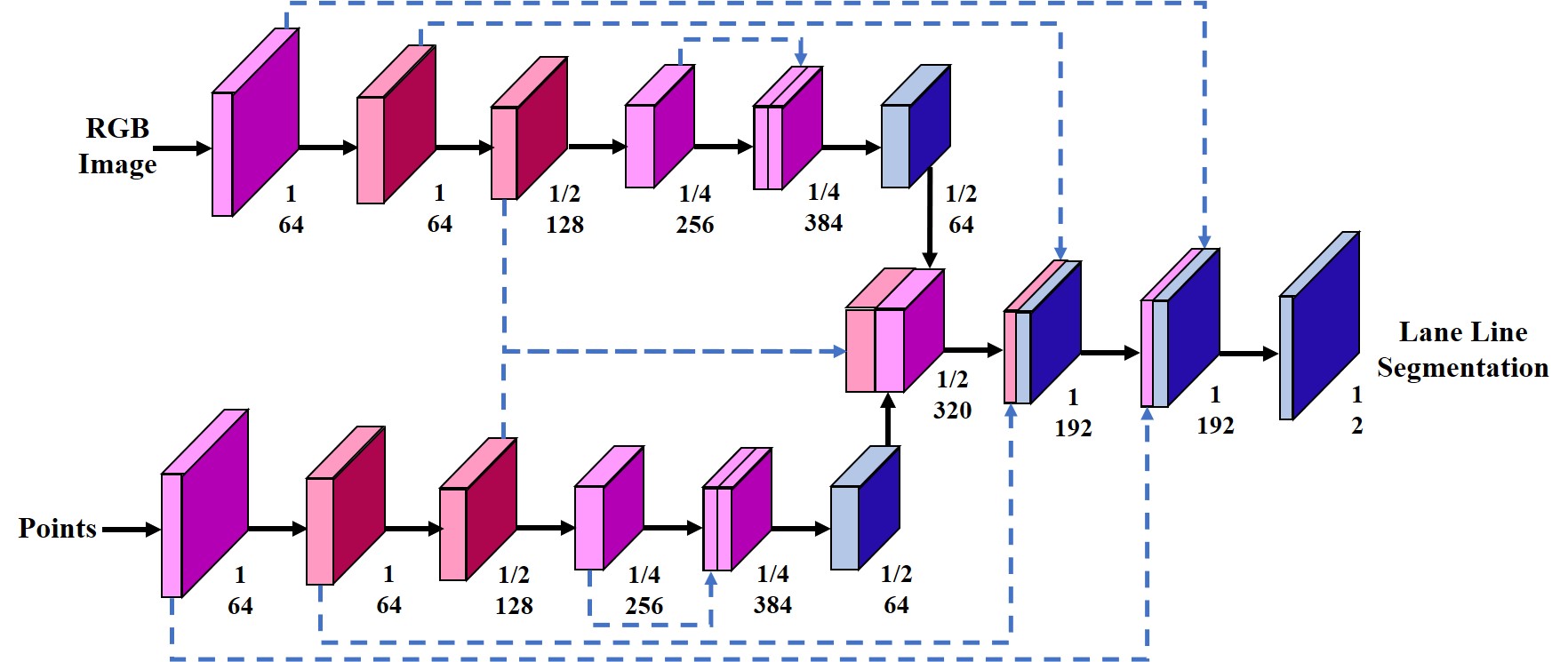}} \\
   \includegraphics[width=1\textwidth]{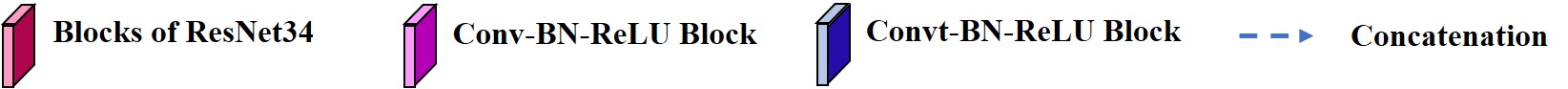}
  \end{tabular}
  \caption{Featue Fusion at Three Stages: the top model (V2\&V3) fuses at the early stage; the middle one (V4) fuses at the middle stage; and the bottom model (V5) conducts fusion in the decoder. Among these model, V2 uses sparse point clouds and V3, V4 and V5 uses completed point clouds. All the three fusion belongs to the 'middle fusion' in \citep{feng2019deep}.}
  \label{Featurefusion}
\end{figure*}

\subsection{Point Cloud Completion}
\par The sparse reflectance is obtained by projecting the font-view point clouds onto the camera imaging plane to align with the RGB images. But the ratio between point clouds and images are around 1.5\% in KITTI and 0.4\% in A2D2. Therefore, point cloud completion can improve the channel capacity with higher SNR. We apply the k-NN search for interpolation: search 3 nearest points for each blank pixel and count the weighted average based on the normalized pixel distance as the result. To better decrease noise from reflectance attenuation and utilize the height information as additional filter condition, we stack the height and distance value on the reflectance to compose pseudo 3-channel images.

\subsection{Fusion at Different Stages}
To validate the optimal fusion stage in network, we build models with the early, middle and late fusion as shown in Fig.\ref{Featurefusion}. In the point clouds pipeline, we use convolutional layers rather than ResNet blocks to achieve the capacity allocation in V4 and V5. Features from two modalities will be concatenated to double channels. Though adaptive bandwidth fusion as mentioned above would perform better, we apply convolution after concatenation for lightweight models.

\subsection{Multi-task Learning}
\begin{figure*}[htp!]
\centering
  \begin{tabular}{@{}c@{}}
  \subfloat[The Multi-task Block And The Adaptive Fusion Block]{%
    \includegraphics[width=1\textwidth]{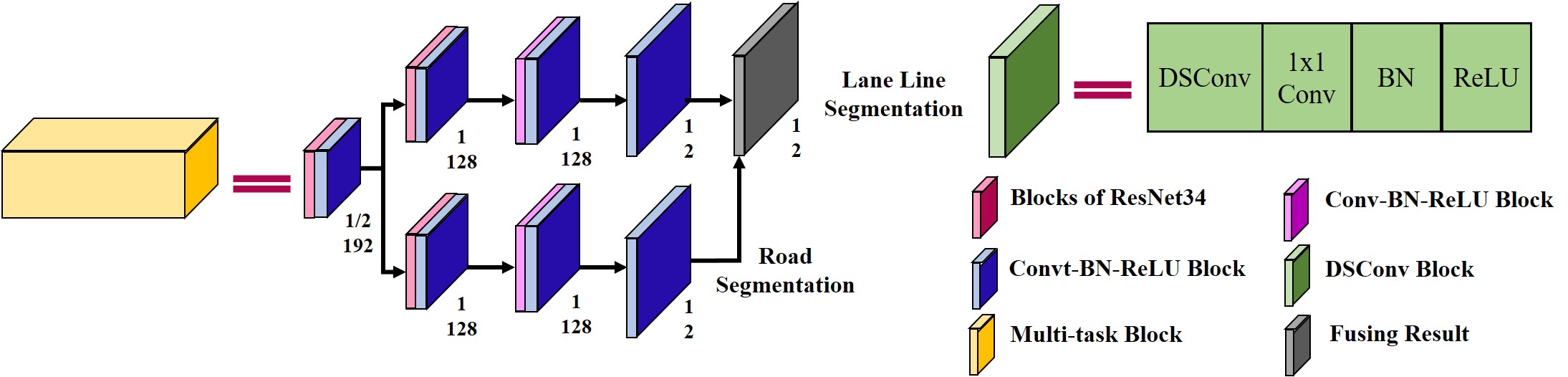}}\\
  \subfloat[The Final Model (V6) with Early \& Middle Fusion]{%
    \includegraphics[width=1\textwidth]{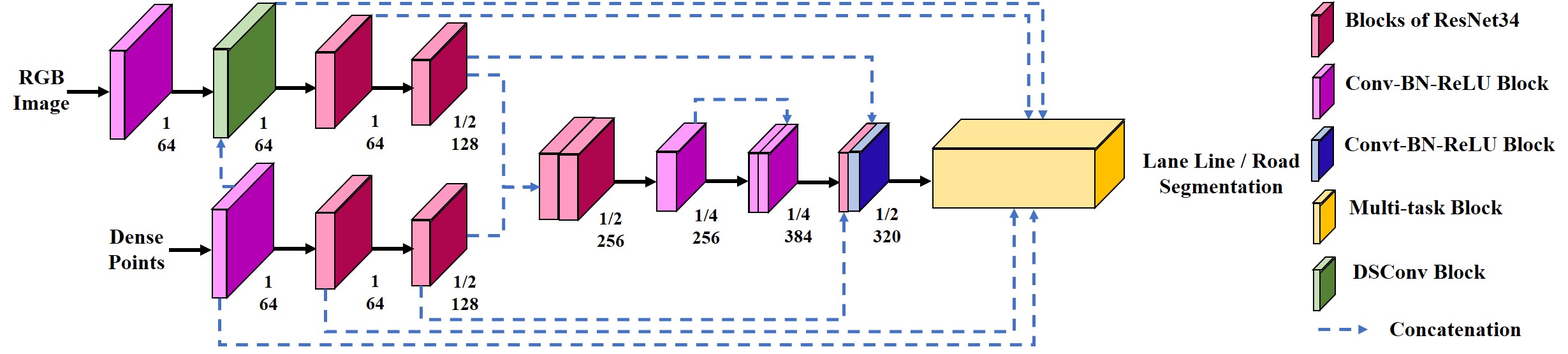}}
  \end{tabular}
  \caption{The Multi-task Block And The Adaptive Fusion Block (DSConv): the gray layer combines the results from two branches.}
  \label{Final}
\end{figure*}

Road segmentation is considered efficient to extract priori knowledge and gain more capacity for lane line. Typically, lane lines exist in the driving area, thus filtering the background in images and point clouds can increase the SNR. Moreover, adding a relevant task equals to adding the code length, which means larger feature subspace. Apart from the capacity, with the Theorem.~\ref{th5}, longer codes can also improve the error correction ability. However, the interference among sub-tasks will slow the convergence rate in training. As shown in the Fig.\ref{Final}, we design a two branch decoder block to conduct multi-task learning, which can be easily extended to other tasks. The block utilizes the last three decoder blocks in each branch, and combine the result to revise the prediction of the main task, lane segmentation. The output of lane is:
\begin{equation}
P(X\in lane|road) = P(X\in lane) \times \{k+(1-k)\times P(X\in road)\}
\label{Twobranch}
\end{equation}
where $k$ is a trainable parameter and the equation helps to decide how much should the lane line prediction relies on road. According to the performance in V3-V5, we add the multi-task block for V3, V4 to build V3r, V4r.

\subsection{Adaptive And Multi-stage Fusion}
We apply a fusion block for the adaptive fusion over multimodal feature channels. It contains a depthwise convolutional layer, a 1x1 convolutional layer, a batch normalization and an activation layer. Therefore, important features will dominate training after the 1x1 convolution. We embed these methods in the V3r model to build V3r+. 
Besides, the final version (V6) are implemented with fusion at multiple stages based on the performance of previous models. Actually, V6 fuses at the early and middle stages simultaneously. This common X-shape architecture can further enlarge the capacity and join the features of V3 and V4.

\section{Experiment}

\subsection{Dataset Preparation}

\begin{figure}[htb]
\centering
  \begin{tabular}{@{}c@{}c@{}c@{}c@{}}
      \subfloat[RGB image]{%
      \begin{tabular}{@{}c@{}}
      \includegraphics[width=0.24\textwidth]{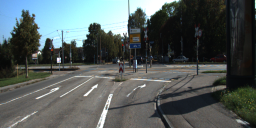}\\
      \includegraphics[width=0.24\textwidth]{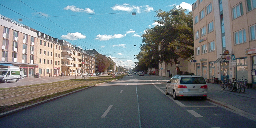}
      \end{tabular}
    } &
      \subfloat[Point Clouds]{%
      \begin{tabular}{@{}c@{}}
      \includegraphics[width=0.24\textwidth]{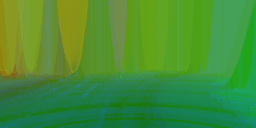}\\
      \includegraphics[width=0.24\textwidth]{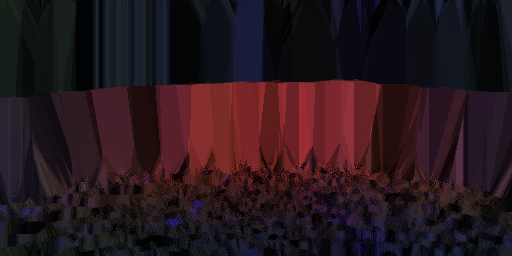}
      \end{tabular}
    } &
      \subfloat[Lane Label]{%
      \begin{tabular}{@{}c@{}}
      \includegraphics[width=0.24\textwidth]{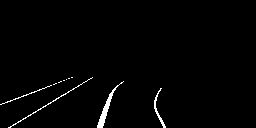}\\
      \includegraphics[width=0.24\textwidth]{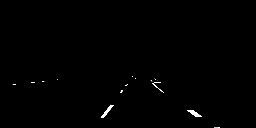}
      \end{tabular}
    } &
      \subfloat[Road Label]{%
      \begin{tabular}{@{}c@{}}
      \includegraphics[width=0.24\textwidth]{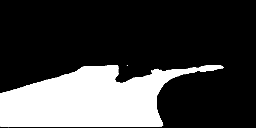}\\
      \includegraphics[width=0.24\textwidth]{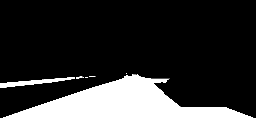}
      \end{tabular}
    }
  \end{tabular}
    \caption{Some Examples of The KITTI (top row) And The A2D2 Dataset (bottom row): the presented point clouds are the composed output of kNN completion.}
  \label{dataset}
\end{figure}

\paragraph{Dataset Overview} To evaluate our models, we select pictures by ignoring the roads with intersections or without forward lines. Finally, we get 383 pairs of data  from the KITTI road detection track\cite{Fritsch2013ITSC}, and 788 pairs from the A2D2 dataset\cite{geyer2020a2d2}. We use 60\% of data as the training set, 10\% for validation and rest for testing. As shown in the Fig.\ref{dataset}, all images contain parallel lane lines and some of them have horizontal lines that do not count in our experiments. In addition, the image resolution is 1242$\times$375 in KITTI, and 1920$\times$1208 in A2D2. KITTI uses a 64-line Velodyne to generate point clouds, but A2D2 combines one 8-line and two 16-line LiDARs. The difference in LiDARs causes the gap in performance. For this reason, the rest comparison will be done on the KITTI rather than A2D2.

\paragraph{Line Annotation} To apply semantic segmentation on the KITTI dataset, we add pixel-level lane line annotation to it.  Comparing with the apolloscape dataset\cite{huang2018apolloscape} and the TuSimple dataset, we filter out confusing lane markings like markings on the sidewalk. Labeled lines are supposed to be not only parallel to the driving direction but also on the driving area. The challenge in KITTI are those complex scenes, including overexposure, darkness, rural area, railway along the road and multiple lines. For more accurate feature extraction, we carefully label those lines whatever in the shadow or under overexposure. The intervals between dash lines are excluded if they are at the ends of line segments. To reduce noise in the annotation, we do not estimate any markings behind obstacles like vehicles and poles on the roadside. Different from KITTI, A2D2 provides similar lane line labels but they ignore the intervals in dash lines.

\paragraph{Data Augmentation} To alleviate the overfitting problem on small dataset like KITTI, we augmented the dataset with geometric transform like perspective translation, random rotation and flipping; pixel-level augmentation like brightness and contrast adjustment; target interference like gaussian noise, random cropping and random erasing part of the lane lines. All augmentation methods were executed on the images except the geometric translation.

\paragraph{Data Preprocess}
To intergrate LiDAR point clouds and RGB images in the same network, projection and value normalization are essential in preprocess. To project the point clouds onto the image plane, given a point $P_v=(x_v,y_v,z_v)^T$, we calculate:
\begin{equation}
P_v'=K_v[R_v|T_v]P_v
\end{equation}
where $K_v,R_v,T_v$ refer to the camera calibration matrix, rotation matrix and translation matrix. Then the projected front-view point cloud reflectance map will be cropped to the same size of RGB images which is 128*256. After that, the value of both reflectance map and the RGB images will be normalize to [0,1] interval.

\subsection{Implementation}
\paragraph{Optimization}
All models are trained with the ADAM optimizer with an cyclical decayed learning rate $lr$:
\begin{equation}
lr = 2^{\left \lfloor epoch/50 \right \rfloor} \times 0.8^{\left \lfloor epoch/10 \right \rfloor} \times lr_0, {\ }lr_0 = 0.0001
\end{equation}
\paragraph{Adaptive Weighted Loss} It is common to use weights in classification, but most methods rely on priori knowledge of the dataset or adjusting parameters during training. Inspired by \cite{he2017adaptive}, we use the $torch.nn.NLLLoss2d$ in PyTorch and set the weight as the inverse ratio of two class in the last prediction. However, sometimes a branch will dominate training and cause unexpected weights, like training with multi-task. For better convergence, we set the weights as (0.5,0.5) in the first 20 epochs.
\paragraph{Hardware and Software}
Our models are trained on a single GPU, GTX 1080Ti with 11G RAM and E5-2678v3 CPU. Besides, we use the following software setup: Ubuntu 16.04 64 bit Operating System, Python 3.6, gcc5.4.0, PyTorch 1.10 with CUDA 9.0 hardware acceleration.

\begin{figure*}[!t]
\centering
\includegraphics[width=1\textwidth]{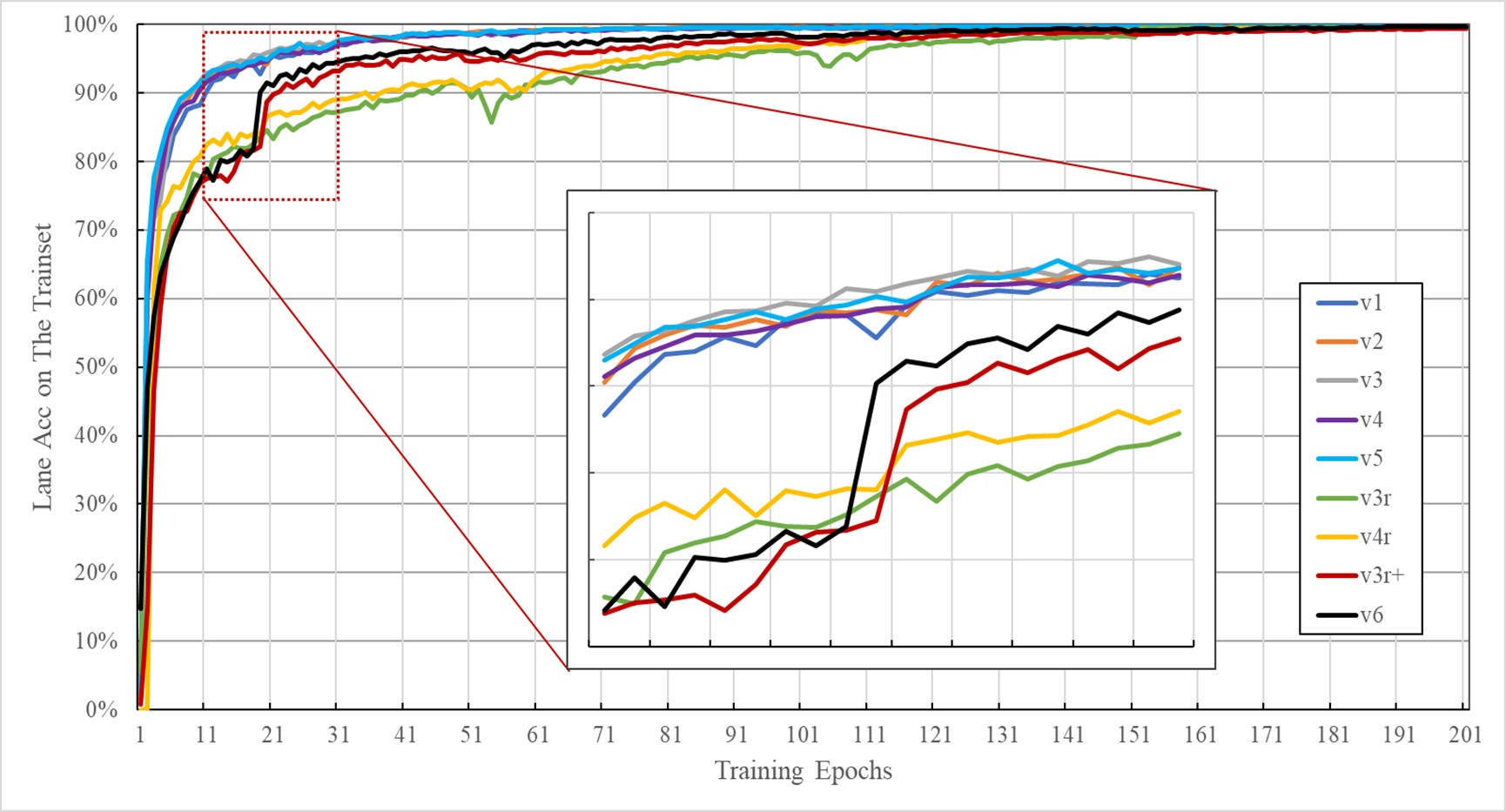}
\caption{The Convergence Process in Training: the curves show the accuracy for lane on the training set. V1: the baseline; V2-V5: feature fusion at early, middle, late stages; V3r,V4r: multi-task training version for V3,V4; V3r+: refined version of V3r by adding adaptive blocks; V6: the full and final version.}
\label{convergence}
\end{figure*}

\subsection{Evaluation}
\subsubsection{Metrics}
We foucs more on the recall of lane line and compute it as the $lane{\ } accuracy$ $(LAcc)$. We also consider the $F2$-$score$ to balance in case the network overfits any class, and count the mean recall on both class as the $mAcc$. In the road segmentation, we apply the same metrics:
\begin{equation}
precise = \frac{TP}{TP+FP},{\ }LAcc=recall=\frac{TP}{TP+FN}
\label{Metrics1}
\end{equation}
\begin{equation}
F2=\frac{(1+2^2)\times precision\times recall}{(2^2)\times precision+recall}
\end{equation}
\begin{equation}
Acc = \frac{TP+TN}{TP+TN+FP+FN}
\label{Metrics2}
\end{equation}
\begin{equation}
mAcc = (\frac{TP}{TP+FN}+\frac{TN}{TN+FP})/2
\end{equation}

In Eq.~\ref{Metrics1}, $TP,TN,FP,FN$ refers to the True Positive, True Negative, False Positive and False Negative.

\subsubsection{Results}
We compare our models (V1-V6,V3r,V4r,V3+) over the selected subset of the KITTI and A2D2 datasets with two leading models, SCNN\cite{pan2018spatial} and LaneNet\cite{neven2018towards}. All models start training from scratch, except that SCNN and LaneNet load the pretrained VGG-16 weights to accelerate learning. To be fair, we train SCNN and LaneNet for 10000 iterations (equal to 175 epochs, for they have stopped optimization after 5000 iteration) and our models for 200 epochs due to the computing resource limitation. Thus, their performance can be better after well fine-tuning or more training. The training record of our models on the KITTI dataset is shown in Fig.~\ref{convergence}. All models reach the same performance after around 150 epochs without significant gap for different fusion stages, but adding the multi-task learning will slow the fitting due to the message conflict in channels. We also notice that V3+ and V6 accelerate after 20 epochs that verifies the influence of adaptive fusion in capturing important features.
\begin{table}[t!]
\centering    
\addtolength{\leftskip} {-3cm}
\addtolength{\rightskip}{-3cm}
\caption{The Results on The KITTI And The A2D2 Datasets: the E/M/L denotes the early/middle/late fusion.}
\renewcommand\arraystretch{1}
\begin{tabular}{l|c|c|p{0.75cm}p{0.75cm}p{0.75cm}p{0.75cm}|p{0.75cm}p{0.75cm}p{0.75cm}p{0.75cm}|p{0.75cm}}
\hline
{ }& { }& { } & \multicolumn{4}{c|}{{\textbf{KITTI(383)}}}                                                                                                  & \multicolumn{4}{c|}{{\textbf{A2D2(788)}}}                                                                                                   & { }                               \\ \cline{4-11}
\multirow{-2}{*}{{\textbf{Model}}} & \multirow{-2}{*}{{\textbf{Fuse}}} & \multirow{-2}{*}{{\begin{tabular}[c]{@{}l@{}}\textbf{Size}\\ \textbf{(M)}\end{tabular}}} & {\textbf{LAcc}} & {\textbf{Acc}} & {\textbf{mAcc}}    & {\textbf{F2}}   & {\textbf{LAcc}} & {\textbf{Acc}} & {\textbf{mAcc}}    & {\textbf{F2}}   & \multirow{-2}{*}{{\textbf{FPS}}} \\ \hline
{lanenet}                          & {/}                              & {556}                            & {70.97}           & {95.78}      & {83.46}          & {33.7}          & {78.83}           & {97.30}      & {88.12}          & {41.5}          & {69.1}                           \\
{scnn}                             & {/}                              & {556}                            & {84.68}           & {98.29}      & {91.55}          & {58.4}          & {80.93}           & {96.06}      & {85.54}          & {34.2}          & {14.4}                           \\ \hline
{V1}                               & {/}                              & {25}                             & {82.38}           & {98.71}      & {90.63}          & {62.2}          & {82.55}           & {98.43}      & {90.54}          & {54.3}          & {\textbf{73.3}}                  \\
{V2}                               & {E}                             & {25}                             & {82.50}           & {98.70}      & {90.68}          & {62.0}          & {81.80}           & {98.68}      & {90.29}          & {57.4}          & {68.1}                           \\
{V3}                            & {E}                             & {25}                             & {84.90}           & {98.73}      & {91.89}          & {\textbf{64.0}}  & {81.55}           & {98.79}      & {90.22}          & {59.6}          & {68.9}                           \\
{V4}                            & {M}                               & {31}                             & {84.46}           & {98.72}      & {91.67}          & {63.5}          & {79.85}           & {98.80}      & {89.38}          & {58.6}          & {\textbf{72.4}}                  \\
{V5}                            & {L}                              & {31}                             & {82.66}           & {98.77}      & {90.80}          & {63.6}          & {70.28}           & {99.10}      & {84.79}          & {57.5}          & {55.0}                           \\
{V3r}                            & {E}                             & {27}                             & {\textbf{86.21}}  & {98.64}      & {\textbf{92.49}} & {63.9}          & {\textbf{86.99}}  & {98.02}      & {\textbf{92.54}} & {51.5}          & {63.2}                           \\
{V4r}                            & {M}                            & {35}                             & {83.47}           & {98.74}      & {91.18}          & {63.3}          & {83.32}           & {98.59}      & {90.51}          & {56.4}          & {63.4}                           \\
{V3r+}                            & {E}                             & {27}                             & {\textbf{85.92}}  & {98.67}      & {\textbf{92.36}} & {63.3}          & {84.90}           & {98.69}      & {\textbf{91.74}} & {59.3}          & {62.3}                           \\ \hline
{V6}                       & {E+M}                         & {35}                             & {\textbf{86.48}}  & {98.76}      & {\textbf{92.69}} & {\textbf{64.9}}          & {85.13}           & {98.86}      & {\textbf{92.04}} & {\textbf{61.9}} & {64.9}                           \\ \hline
\end{tabular}
\label{allresult}
\end{table}
\par According to the Table.~\ref{allresult}, most models achieve over 98.7\% accuracy on both datasets. But all F2-score are lower than 65.0, that can be caused by the severe unbalance between lane line and the background (over 98\% are background). False prediction on very few pixels can lead to a huge variation in recall and precise, which can be solved by using precise post-processing, let alone the deviation caused by annotation. As for LAcc, we found that the earlier it fuse, the better it performs. Besides, V3r and V4r get higher scores than V3 and V4, indicating the benefits of multi-task learning. By comparing the results on KITTI and A2D2, we found that the point clouds completion even weaken the model. Since it can be caused by the low quality of A2D2 LiDAR for lane line segmentation. The model V6, which is built according to the results of V3-V5. It overtakes other versions in almost all tests.
\par Finally, we count the time cost in segmentation network forward as the model speed. All our models except V5 can reach at least 62 FPS with less than 4G memory on GPU. As shown in the table, our lightweight models can approach the cutting-edge performance on both KITTI and A2D2 datasets and meet the requirements of autonomous driving. Notice that our models are built on a basic backbone, and they are supposed to be better with more optimal achitecture like knowledge distillation\cite{hinton2015distilling} and multi-objective particle swarm optimization\cite{chouikhi2019bi} in the future work. To further clarify what contributes to the performance, we conduct more experiments on the KITTI and augmented KITTI datasets with our models in the following ablation study.

\section{Ablation Study}
In these section, we further conduct quantitative analysis on the proposed strategy. First, we discuss the contribution from different factors in fusion from the viewpoint of channel, then compare the robustness in case that the models lose one modality in practical usage, which also verifies our inference.
\par
\begin{sidewaystable}[htp!]
\caption{The Comparison of Different Fusion Strategy: dense denotes the point clouds completion, and adap is the adaptive block.}
\renewcommand\arraystretch{0.9}
\begin{tabular}{p{1.2cm}|p{1.1cm}|p{0.8cm}p{0.8cm}p{0.6cm}p{0.6cm}p{0.7cm}p{0.9cm}|p{1.05cm}p{1.05cm}p{1.05cm}|p{1.05cm}p{1.05cm}p{1.05cm}}
\hline
{}& {} & \multicolumn{6}{c|}{{ \textbf{Fusion}}}                                                                                                                                                                                                                                                     & \multicolumn{3}{c|}{{ \textbf{KITTI}}}                                                               & \multicolumn{3}{c}{{ \textbf{KITTI (Aug)}}}                                                             \\ \cline{3-14}
\multirow{-2}{*}{\textbf{Data}} & \multirow{-2}{*}{\textbf{Model}}& {\textbf{dense}}       & {\textbf{early}}            & {\textbf{mid}}           & {\textbf{late}}             & {\textbf{road}}       & {\textbf{adap}}           & {\textbf{LAcc}} & {\textbf{mAcc}}    & {\textbf{F2}}  & {\textbf{LAcc}} & {\textbf{mAcc}}   & {\textbf{F2}} \\ \hline
{} & {V1}  & {}   & {}   & {}   & {}   & {}   & {}   & { 82.38}     & { 90.63}    & { 62.2}   & { 82.55}     & { 90.54}   & { 54.3}      \\ \cline{2-14}
{} & { V2}  & {}   & { \checkmark} & {}   & {}   & {}   & {}   & { +0.12}      & { +0.05}     & { $-$0.2}   & { $-$1.86}     & { $-$1.21}   & { +3.7} \\
{} & { V3}     & { \checkmark} & {\checkmark} & {}   & {}   & {}   & {}   & { +2.52}      & { +1.26}     & { \textbf{+1.9}} & { $-$1.00}     & { $-$0.32}   & { +5.3}    \\
{} & { V4}     & { \checkmark} & {}   & { \checkmark} & {}   & {}   & {}   & { +2.08}      & { +1.04}     & { +1.3}    & { $-$2.70}     & { $-$1.16}   & { +4.3} \\
{} & { V5}     & { \checkmark} & {}   & {}   & { \checkmark} & { \textbf{}}& {}   & { +0.28}      & { +0.17}     & { +1.4}    & { $-$12.27}    & { $-$5.75}   & { +3.2} \\
{} & { V3r}     & { \checkmark} & { \checkmark} & {}   & {}   & { \checkmark}   & {}   & { \textbf{+3.83}}   & { \textbf{+1.86}}  & { +1.7} & { \textbf{+4.44}}   & { \textbf{+2.00}} & { $-$2.8}      \\
{} & { V4r}     & { \checkmark} & {}   & { \checkmark} & {}   & { \checkmark} & {}   & { +1.09}      & { +0.55}     & { +1.1}    & { +0.77}      & { $-$0.03}   & { +2.1} \\
{} & { v3r+}     & { \checkmark} & { \checkmark} & {} & { \textbf{}}& { \checkmark} & { \checkmark} & { \textbf{+3.54}}   & { \textbf{+1.73}}  & { +1.1}    & { +2.35}      & { +1.20}    & { +5.0}      \\ \cline{2-14}
\multirow{-9}{*}{{\textbf{KITTI}}} & { V6}      & { \checkmark} & { \checkmark} & { \checkmark} & {}   & { \checkmark} & { \checkmark} & { \textbf{+4.10}}   & { \textbf{+2.06}}  & { \textbf{+2.7}} & { +2.58}      & { \textbf{+1.50}} & { \textbf{+7.6}}    \\ \hline
{} & { V1}  & {}   & { \textbf{}}& {}   & { \textbf{}}& {}   & {}   & { 85.29}     & { 92.14}    & { 66.0}   & { 79.45}     & { 85.84}   & { 56.4}      \\ \cline{2-14}
{} & { V2}  & {}   & { \checkmark} & {}   & { \textbf{}}& {}   & {}   & { $-$2.28}     & { $-$1.04} & { +1.3}    & { $-$1.10}     & { $-$0.25}   & { $-$0.3}      \\
{} & { V3}     & { \checkmark} & { \checkmark} & {}   & {}   & {}   & {}   & { $-$0.05}     & { $-$0.25}    & { \textbf{+3.7}} & { +3.15}      & { +3.09} & { \textbf{+6.1}}    \\
{} & { V4}     & { \checkmark} & {}   & { \checkmark} & {}   & {}   & {}   & { +0.41}      & { +0.27}     & { +2.1}    & { +2.04}      & { +0.87}    & { +3.0} \\
{} & { V5}     & { \checkmark} & {}   & {}   & { \checkmark} & {}   & {}   & { $-$3.74}     & { $-$1.75}    & { +1.3}    & { $-$2.60}     & { $-$0.63}   & { +2.4} \\
{} & { V3r}     & { \checkmark} & { \checkmark} & {}   & {}   & { \checkmark}   & {}   & { \textbf{+3.75}}   & { \textbf{+1.62}}  & { $-$3.6}   & { +5.47}      & { +1.64}    & { $-$8.7}      \\
{} & { V4r}     & { \checkmark} & {}   & { \checkmark} & {}   & { \checkmark} & {} & { \textbf{+3.58}}   & { \textbf{+1.55}}     & { $-$3.1}   & { \textbf{+7.30}}   & { +0.56}    & { $-$0.9}      \\ \cline{2-14}
\multirow{-8}{*}{{\begin{tabular}[c]{@{}l@{}}\textbf{KITTI}\\ \textbf{(Aug)}\end{tabular}}} & { V6}  & { \checkmark} & { \checkmark} & { \checkmark} & {}   & { \checkmark} & { \checkmark} & { \textbf{+4.60}}   & { \textbf{+2.15}}  & { $-$0.2} & { \textbf{+8.51}}   & { \textbf{+6.21}} & { +2.2} \\ \hline
\end{tabular}
\label{ablation}
\end{sidewaystable}

\subsection{Individual Contribution in Fusion}
To explore the influence of source and channel together, we train and test our models on the KITTI and its augmented dataset separately. Their performance in LAcc, mAcc and F2-score are listed in the Table.~\ref{ablation}. Based on the result, V3r, V3r+ and V6 are the best when training with the original KIITI data, and V3r, V4r, V6 lead in the augmented training. However, the fusion models can be even worse in some cases: the V5 is weaken on the augmented dataset, and V2-V5 perform not really well when using KITTI for training and the augmented data for testing. The unstable performance reflects the potential dependence on information source. Apart from this, we assume every factor are linearly independent and quantify their effect by solving the Eq.~\ref{factors} by least squares, finally obtain the result in Table.~\ref{factorresult}. Notice that the value do not indicate the absolute gain or loss in fusion, but the relative contribution. 
\begin{equation}
{
\left[
\begin{array}{ccccccc}
1	&	0	&	1	&	0	&	0	&	0	&	0	\\
1	&	1	&	1	&	0	&	0	&	0	&	0	\\
1	&	1	&	0	&	1	&	0	&	0	&	0	\\
1	&	1	&	0	&	0	&	1	&	0	&	0	\\
1	&	1	&	1	&	0	&	0	&	1	&	0	\\
1	&	1	&	0	&	1	&	0	&	1	&	0	\\
1	&	1	&	1	&	1	&	0	&	1	&	1	
\end{array}
\right ]
}
\times
{
\left[
\begin{array}{c}
LiDAR	\\
dense	\\
early	\\
middle	\\
late	\\
road	\\
adaptive
\end{array}
\right ]
}
=
{
\left[
\begin{array}{l}
Result{\ }	of{\ }	V2	\\
Result{\ }	of{\ }	V3	\\
Result{\ }	of{\ }	V4	\\
Result{\ }	of{\ }	V5	\\
Result{\ }	of{\ }	V3r	\\
Result{\ }	of{\ }	V4r	\\
Result{\ }	of{\ }	V6	
\end{array}
\right ]
}
\label{factors}
\end{equation}

\begin{table}[htp]
\caption{The Quantitive Comparison of Different Fusion Factors: the first letter in a group indicates the training set and the other is the test set. The letter K means the KITTI dataset, and A means the augmented data.}
\renewcommand\arraystretch{1}	

\begin{tabular}{p{1.1cm}|p{0.85cm}p{0.85cm}|p{0.85cm}p{0.85cm}|p{0.85cm}p{0.85cm}|p{0.85cm}p{0.85cm}}
\hline
{}	&	\multicolumn{4}{c|}{{\textbf{LAcc}}}	&	\multicolumn{4}{c}{{\textbf{mAcc}}}	\\	\cline{2-9}														 
\multirow{-2}{*}	&	{\textbf{K-K}}	&	{\textbf{A-K}}	&	{\textbf{K-A}}	&	{\textbf{A-A}}	&	{	\textbf{K-K}}		&	{	\textbf{A-K}}		 &	{	\textbf{K-A}}		&	{	\textbf{A-A}}		\\	\hline
{\textbf{LiDAR}}	&	{$-$1.27}	&	{$-$2.74}	&	{$-$4.64}	&	{$-$2.20}	&	{$-$0.63}	&	{$-$1.21}	&	{$-$2.65}	&	{$-$2.40}	\\	 \hline
{\textbf{dense}}	&	{+2.98}	&	{+2.39}	&	{+1.35}	&	{+3.52}	&	{+1.48}	&	{+0.94}	&	{+1.19}	&	{+3.06}	\\	\hline
{\textbf{early}}	&	{+1.39}	&	{+0.46}	&	{+2.78}	&	{+1.10}	&	{+0.68}	&	{+0.17}	&	{+1.44}	&	{+2.15}	\\	\hline
{\textbf{mid}}	&	{$-$0.20}	&	{+0.60}	&	{+0.10}	&	{+1.46}	&	{$-$0.08}	&	{+0.39}	&	{+0.00}	&	{+0.50}	\\	\hline
{\textbf{late}}	&	{$-$1.42}	&	{$-$3.39}	&	{$-$8.98}	&	{$-$3.92}	&	{$-$0.68}	&	{$-$1.48}	&	{$-$4.29}	&	{$-$1.28}	\\	 \hline
{\textbf{road}}	&	{+0.16}	&	{+3.49}	&	{+4.46}	&	{+3.79}	&	{+0.06}	&	{+1.58}	&	{+1.73}	&	{$-$0.88}	\\	\hline
{\textbf{adap}}	&	{+1.04}	&	{+0.41}	&	{$-$1.46}	&	{+0.85}	&	{+0.55}	&	{+0.29}	&	{$-$0.20}	&	{+3.78}	\\	\hline

\end{tabular}
\label{factorresult}
\end{table}
\par According to the Table.\ref{factorresult}, the fusion network has different performance based on the train/test case. The raw data is a subset of the augmented data, that means the case A-A and A-K have smaller train-test distance, while K-K and K-A have larger gap. Actually, when we focus on the same row, we observe early fusion exceeds the middle fusion when training on the KITTI, but the case will be inverse on the augmented dataset, and the late fusion performs even worse without enough capacity in fusion channel. In addition, the ratio $early/middle$ decreases when the training set changes from K to A, or the test set changes from A to K, however, it increases from A-K to K-A. The contrast is also obvious if we normalize the value in table by $\hat{x}_i=(x_i-x_{min})/\Sigma(x_i-x_{min})$. The main divergence in these train/test cases is the uncertainty, and the results verify our hypothesis about pre-fusion and post-fusion, as well as the impact of source. Our conclusion can be generalized to the situation that part of sensors fail to record accurately or even break down, and we also present the result in the modality lost experiment.
\par Apart from these, we also learn the raw point clouds can harm the network, because the sparse signal causes low SNR and reduce the channel capacity. In contrast, point clouds completion will make up for this deficiency, which is actually the most beneficial factor in fusion. Besides, multi-task learning (road) and adaptive fusion gain more capacity in almost all cases. Especially, the trend of road segmentation is close to the one of middle fusion for they serve similarly to increase the channel capacity. The adaptive fusion performs better when the training set is close to the test set, like K-K and A-A, but the result is more likely to be irregular. Possible reasons include the shortcomings of our simple design, competition with multi-task (they keep around 37\% normalized contribution in all cases) or the over-fitting on the training set, for it only optimize the capacity allocation of the given channel based on the input data. It needs more experiments to acquire the convincing illustration.

\begin{figure*}[htp]
\centering
  \begin{tabular}{@{}c@{ }c@{ }c@{ }c@{ }c@{ }c@{ }c@{ }c@{ }c@{ }c}
    \includegraphics[width=0.155\textwidth]{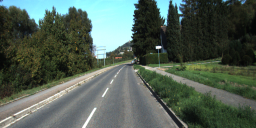} &
    \includegraphics[width=0.155\textwidth]{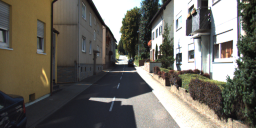} &
    \includegraphics[width=0.155\textwidth]{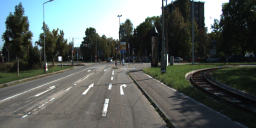} &
    \includegraphics[width=0.155\textwidth]{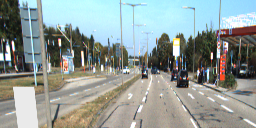} &
    \includegraphics[width=0.155\textwidth]{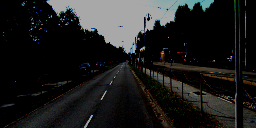} &
    \includegraphics[width=0.155\textwidth]{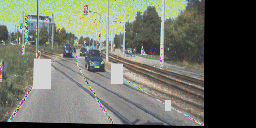}
\\
    \includegraphics[width=0.155\textwidth]{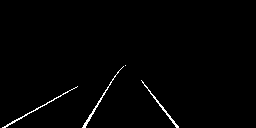} &
    \includegraphics[width=0.155\textwidth]{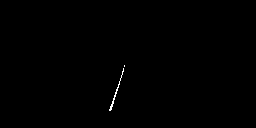} &
    \includegraphics[width=0.155\textwidth]{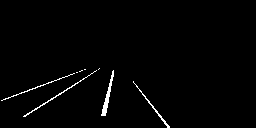} &
    \includegraphics[width=0.155\textwidth]{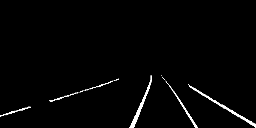} &
    \includegraphics[width=0.155\textwidth]{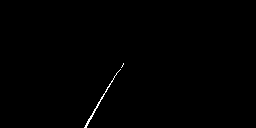} &
    \includegraphics[width=0.155\textwidth]{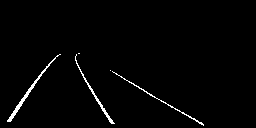}
\\
    \includegraphics[width=0.155\textwidth]{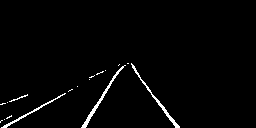} &
    \includegraphics[width=0.155\textwidth]{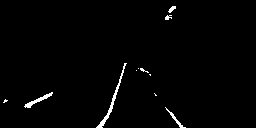} &
    \includegraphics[width=0.155\textwidth]{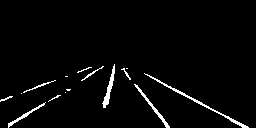} &
    \includegraphics[width=0.155\textwidth]{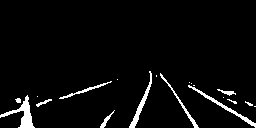} &
    \includegraphics[width=0.155\textwidth]{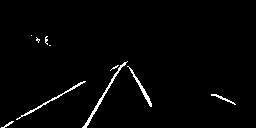} &
    \includegraphics[width=0.155\textwidth]{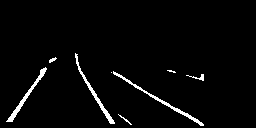}
\\
    \includegraphics[width=0.155\textwidth]{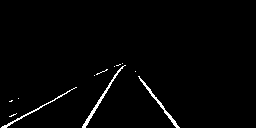} &
    \includegraphics[width=0.155\textwidth]{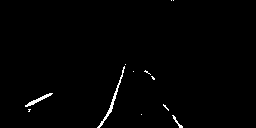} &
    \includegraphics[width=0.155\textwidth]{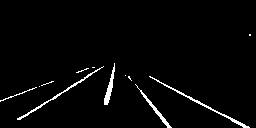} &
    \includegraphics[width=0.155\textwidth]{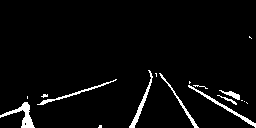} &
    \includegraphics[width=0.155\textwidth]{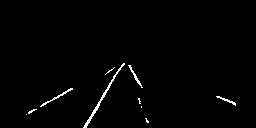} &
    \includegraphics[width=0.155\textwidth]{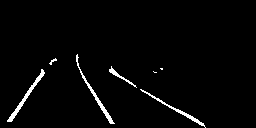}
\\
    \includegraphics[width=0.155\textwidth]{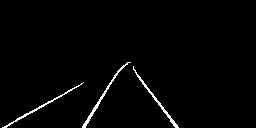} &
    \includegraphics[width=0.155\textwidth]{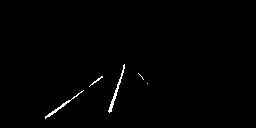} &
    \includegraphics[width=0.155\textwidth]{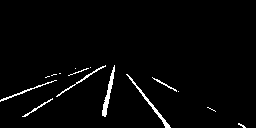} &
    \includegraphics[width=0.155\textwidth]{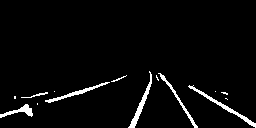} &
    \includegraphics[width=0.155\textwidth]{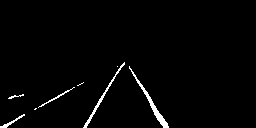} &
    \includegraphics[width=0.155\textwidth]{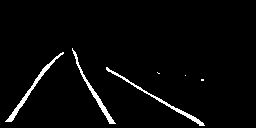}
\\
    \includegraphics[width=0.155\textwidth]{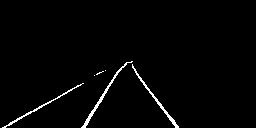} &
    \includegraphics[width=0.155\textwidth]{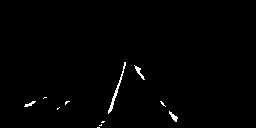} &
    \includegraphics[width=0.155\textwidth]{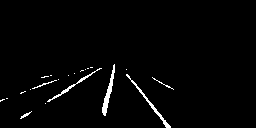} &
    \includegraphics[width=0.155\textwidth]{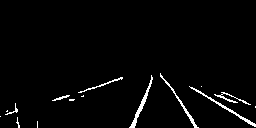} &
    \includegraphics[width=0.155\textwidth]{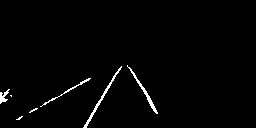} &
    \includegraphics[width=0.155\textwidth]{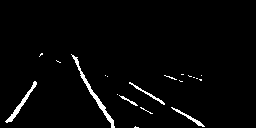}
\\
    \includegraphics[width=0.155\textwidth]{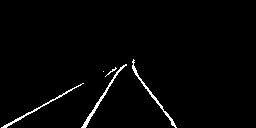} &
    \includegraphics[width=0.155\textwidth]{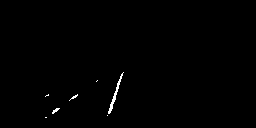} &
    \includegraphics[width=0.155\textwidth]{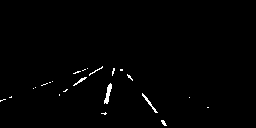} &
    \includegraphics[width=0.155\textwidth]{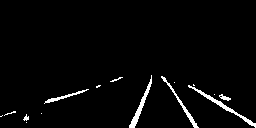} &
    \includegraphics[width=0.155\textwidth]{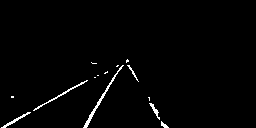} &
    \includegraphics[width=0.155\textwidth]{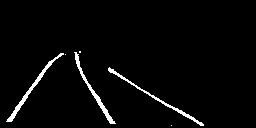}
\\
    \includegraphics[width=0.155\textwidth]{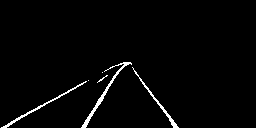} &
    \includegraphics[width=0.155\textwidth]{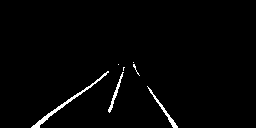} &
    \includegraphics[width=0.155\textwidth]{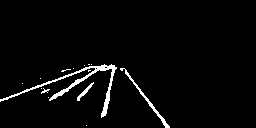} &
    \includegraphics[width=0.155\textwidth]{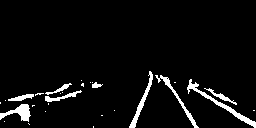} &
    \includegraphics[width=0.155\textwidth]{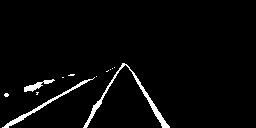} &
    \includegraphics[width=0.155\textwidth]{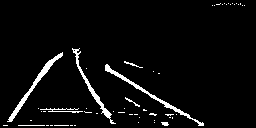}
\\
    \includegraphics[width=0.155\textwidth]{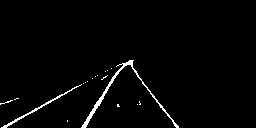} &
    \includegraphics[width=0.155\textwidth]{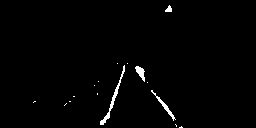} &
    \includegraphics[width=0.155\textwidth]{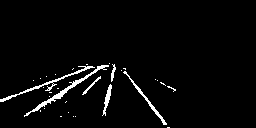} &
    \includegraphics[width=0.155\textwidth]{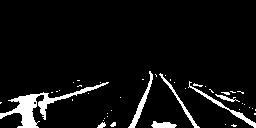} &
    \includegraphics[width=0.155\textwidth]{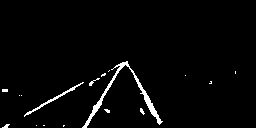} &
    \includegraphics[width=0.155\textwidth]{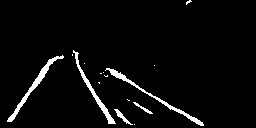}
\\
    \includegraphics[width=0.155\textwidth]{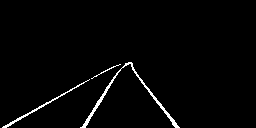} &
    \includegraphics[width=0.155\textwidth]{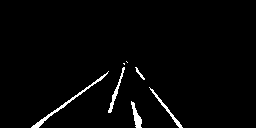} &
    \includegraphics[width=0.155\textwidth]{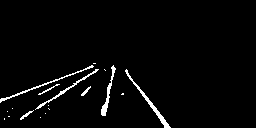} &
    \includegraphics[width=0.155\textwidth]{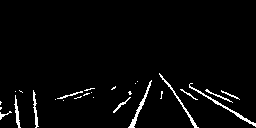} &
    \includegraphics[width=0.155\textwidth]{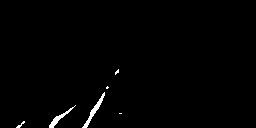} &
    \includegraphics[width=0.155\textwidth]{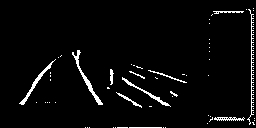}
\\
    \includegraphics[width=0.155\textwidth]{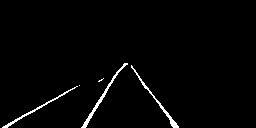} &
    \includegraphics[width=0.155\textwidth]{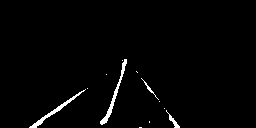} &
    \includegraphics[width=0.155\textwidth]{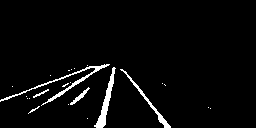} &
    \includegraphics[width=0.155\textwidth]{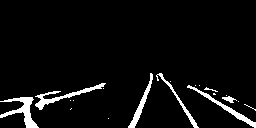} &
    \includegraphics[width=0.155\textwidth]{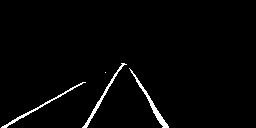} &
    \includegraphics[width=0.155\textwidth]{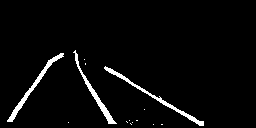}
\\

  \end{tabular}
  \caption{Examples of The Test Result on The KITTI Dataset: the three left columns are tested on the KITTI data and the rest are tested with the augmented data. The rows from top to bottom are: input images, ground truth, output from V1-V5, V3r, V4r, V3r+ and V6.}
  \label{Easyhard}
\end{figure*}

\par Some testing examples are presented in the Fig.~\ref{Easyhard}. The line prediction in the last four rows are more clear than others, because their predicted lines occupy more pixels on the boundaries, which is acceptable in usage and can be refined with extra post-process.

\subsection{Test with Single Modality Lost}
It is unavoidable that error occurs in part of the sensors or data registration, both of which can result in the modality lost in model. Thus, we use zero tensors as the lost data (another data is preserved) to simulate the situation in testing and compare the robustness of models. The weight of single modality in the fused codes (features) can also be deduced according to the Theorem.~\ref{th5}. Moreover, the models are trained with augmented data and tested with KITTI. 
\par As illustrated earlier, early fusion can generalize to other tasks better with more post-fusion, while middle fusion at later stage can fit the data more with deeper individual processes for different modalities. As shown in the Table.~\ref{dropmodal}, V3/V3r surpass V4/V4r with only images, a case never appear in the training set (the augmentation is mainly for images), while the latter models perform better with only point clouds. In general, in our experiment setting, middle fusion is more stable in the case of modality lost. Besides, road segmentation as multi-task learning can extend the overall code length and capacity, thus enhancing error correction for both data. 
\begin{table}[htp]
\caption{Experiments with Single Modal Data Lost: the values in the last group (Random) are the average of the two modality lost cases.}
\renewcommand\arraystretch{1}	
\begin{tabular}{p{1cm}|p{0.9cm}p{0.9cm}|p{0.85cm}p{0.86cm}|p{0.85cm}p{0.85cm}|p{0.85cm}p{0.85cm}}
\hline
{}                               & \multicolumn{2}{c|}{{\textbf{Image+Points}}}             & \multicolumn{2}{c|}{{\textbf{Only Image}}}               & \multicolumn{2}{c|}{{ \textbf{Only Points}}}              & \multicolumn{2}{c}{{\textbf{Random}}}                  \\ \cline{2-9}
\multirow{-2}{*}{\textbf{Model}}& {\textbf{LAcc}}    & {\textbf{mAcc}} & {\textbf{LAcc}}    & {\textbf{mAcc}}  & {\textbf{LAcc}}    & {\textbf{mAcc}}  & {\textbf{LAcc}}    & {\textbf{mAcc}}  \\ \hline
{V3}                               & {85.24}          & {91.89}      & {56.74}          & {77.76}      & {20.29}          & {60.15}      & {38.52}          & {68.98}      \\ \hline
{V4}                               & {85.70}          & {91.67}      & {51.70}          & {74.04}      & {48.88}          & {74.44}      & {50.29}          & {74.24}      \\ \hline
{V3r}                              & {\textbf{89.04}} & {92.49}      & {63.09}          & {80.17}      & {37.36}          & {66.72}      & {50.23}          & {73.44}      \\ \hline
{V4r}                              & {\textbf{88.87}} & {91.18}      & {55.54}          & {74.86}      & {\textbf{67.88}} & {83.47}      & {61.71}          & {79.17}      \\ \hline
{V6}                             & {\textbf{89.89}} & {92.69}      & {\textbf{75.77}} & {85.89}      & {50.34}          & {74.51}      & {\textbf{63.06}} & {80.21}      \\ \hline
\end{tabular}
\label{dropmodal}
\end{table}
\par Besides, V6 performs better than V3/V3r, but utilize less from point clouds than V4r. Comparing with V3/V3r, V6 not only adds the post-fusion stage by parallel coding after the first fusion, that makes it raise the SNR in channel, but also extends the code length and capacity by adding external information of the point clouds in the middle stage, and achieve better coding on the source. In other words, V6 acquire better post-fusion and distortion/redundancy balance than V3/V3r. Besides, the two-stage fusion force the channel to allocate larger weight for point clouds than V3/V3r, that makes the results more balance in two cases. However, for the different architecture V4r, the X-shape fusion makes the transmission in point clouds pipeline influenced by the images, finally reducing the capacity in it. In conclusion, V6 or other multi-stage fusion architecture is essentially moving the balance in capacity allocation. The result also shows that the global optimization in channels is not always accompanied by the local optimization. 

\par Moreover, according to the Theorem.~\ref{th5}, the modality-lost test also reflect the error correction ability of single-modal data, and further implies the actual weight in the fusion. Specifically, import modality would occupy large weight in the fused code, and tend to require longer code for correction in case the data is lost. Namely, the code performs worse without it. Based on the result in the Table.~\ref{factorresult}, V3/V4/V3r/V6 are better with only images, which means images contribute more in these lane line segmentation models. Then we can apply corresponding capacity allocation in the channel.

\section{Conclusions}
In this paper, we propose a novel camera-LiDAR fusion model for lane line segmentation. By leveraging the information from different sensors, our model can achieve the cutting-edge performance on the KITTI benchmark without pretraining or post-process. Furthermore, we formulate the multimodal network in the framework of channel, and utilize Shannon's theory to reveal the fusion mechanism. Based on the analysis, we suggest approaching three balance about source and channel capacity. We also provide practical methods to compare the contribution of different modalities, methods and fusion stage, which will lead to the optimal fusion structure. Experiments have shown the benefits from our information-driven fusion strategy and architecture. In the future, we will continue the work on quantification like uncertainty estimation, and utilize more about joint coding model in our network. Besides, our work is supposed to integrated well in other areas like detection and localization, which would be helpful for the future development in deep multimodal learning and autonomous driving.

\section*{Acknowledgements}
This work was supported by the National High Technology Research and Development Program of China under Grant No. 2018YFE0204300, Beijing Municipal Science and Technology Commission special major under Grant No. D171100005017002, National Natural Science Foundation of China under Grant No. U1664263, the Grant from the Institute Guo Qiang, Tsinghua University.

\section*{References}
\bibliographystyle{unsrt}
\bibliography{mybibfile}

\end{document}